\documentclass{article}

\usepackage[final]{corl_2017} 
\usepackage{graphicx}
\usepackage{url}
\usepackage{graphics}
\usepackage{caption}
\usepackage{subcaption}
\usepackage{algorithmic}
\usepackage{setspace}
\usepackage{epsfig}
\usepackage{amsfonts}
\usepackage{amsmath}
\usepackage{multirow}
\usepackage{todonotes}
\usepackage{pbox}
\usepackage{booktabs}

\usepackage{color}
\definecolor{CommentRed}{rgb}{0.7,0,0}
\definecolor{CommentBlue}{rgb}{0,0,0.7}
\definecolor{CommentGreen}{rgb}{0,0.7,0}
\definecolor{CommentGreenBlue}{rgb}{0,0.7,0.7}

\newcommand{\method}{Mutual Alignment Transfer Learning}
\newcommand{\meth}{MATL}

\usepackage[ruled]{algorithm2e}
\usepackage{enumitem}

\usepackage{pdfrender}
\DeclareRobustCommand*{\pmbb}[1]{%
  \textpdfrender{
    TextRenderingMode=Stroke,
    LineWidth=.1pt,
  }{#1}%
}

\title{\method}

\author{
  Markus Wulfmeier\thanks{Work done as visiting scholar at Berkeley Artificial Intelligence Lab (BAIR), UC Berkeley}\\
  Oxford Robotics Institute \\
  University of Oxford \\
  \texttt{markusw@robots.ox.ac.uk} \\
   \And
   Ingmar Posner \\
   Oxford Robotics Institute \\
   University of Oxford \\
   \texttt{ingmar@robots.ox.ac.uk} \\
   \AND
   Pieter Abbeel \\
   Berkeley AI Research (BAIR)\\
  University of California Berkeley\\
  OpenAI, San Francisco\\
  \texttt{pabbeel@cs.berkeley.edu} 
}

\begin{document}
\maketitle


\begin{abstract}
Training robots for operation in the real world is a complex, time consuming and potentially expensive task. Despite significant success of reinforcement learning in games and simulations, research in real robot applications has not been able to match similar progress. While sample complexity can be reduced by training policies in simulation, such policies can perform sub-optimally on the real platform given imperfect calibration of model dynamics. We present an approach -- supplemental to fine tuning on the real robot -- to further benefit from parallel access to a simulator during training and reduce sample requirements on the real robot. The developed approach harnesses auxiliary rewards to guide the exploration for the real world agent based on the proficiency of the agent in simulation and vice versa. In this context, we demonstrate empirically that the reciprocal alignment for both agents provides further benefit as the agent in simulation can adjust to optimize its behaviour for states commonly visited by the real-world agent.
\end{abstract}



\section{Introduction}

Recent work in reinforcement learning has led to significant successes such as outperforming humans on a multitude of computer games \citep{mnih2015human,VanSeijen2017} and surpassing the best human players in the games of Chess \cite{campbell2002deep} and Go \citep{Silver2016}.
The principal commonality between these settings is the availability of virtually unlimited training data as these systems can be trained in parallel and significantly faster than real-time, real-world executions. 

However, training agents for operation in the real world presents a significant challenge to the reinforcement learning paradigm as it is constrained to learn from comparatively expensive and slow task executions. 
In addition, limits in the perception system and the complexity of manually providing informative real world rewards for high complexity tasks often result in only sparse and uninformative feedback being available, further increasing sampling requirements. As a result, many tasks which involve physical interaction with the real world and are simple for humans, present insurmountable challenges for robots \cite{moravec1988mind}.

While there has been significant progress towards fast and reliable simulators  \citep{dartsim,todorov2012mujoco,coumans2015bullet}, they do not represent exact replications of the platforms and environments we intend to emulate. 
Systematic model discrepancies commonly prevent us from directly porting policies from simulation to the real platform.

The principal differences between simulation and real world are based on the type of system observations as well as discrepancies between system dynamics. 
Recent developments aim at designing visually more similar environments \citep{nvidiaisaac,airsim2017fsr} and current research targets adapting policies to be invariant with respect to differences between the observation spaces of simulator and real platform \cite{stadie2017,tobin2017domain,Long0J16a, Ganin2016}.

Fine tuning pretrained policies from simulation on the real platform is a straightforward approach to address discrepancies between both systems' dynamics. However, as policies trained via reinforcement learning will learn to exploit the specific characteristics of a system -- optimizing for mastery instead of generality -- a policy can overfit to the simulation. The resulting initialization can prove unfavourable to random initializations for further training on the real platform as it might inhibit exploration and lead to the optimization process getting stuck in local optima. A phenomenon which we demonstrate in the experiments in Section \ref{sec:frameworks}. On the other hand, in cases where fine tuning improves performance it can be straightforwardly combined with the presented approach as described in Section \ref{sec:exps}.

\begin{figure}[h]
\centering
\includegraphics[width=.5\textwidth]{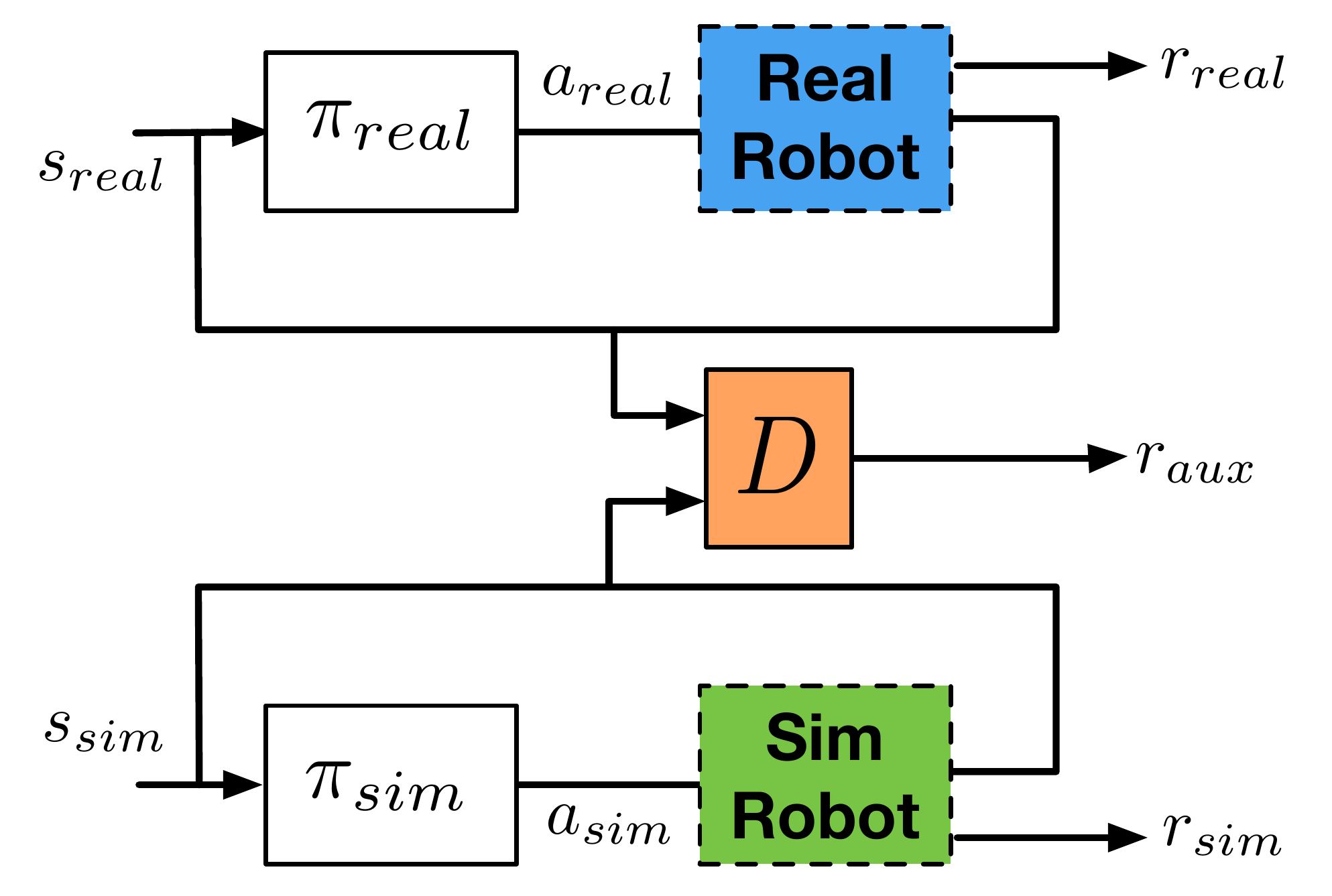}
\caption{Simplified schema for \method. Both systems are trained to not only maximize their respective environment rewards but also auxiliary alignment rewards that encourages both systems to occupy similar distributions over visited states. Furthermore, the simulation policy can be trained orders of magnitude faster than the real platform solely based on its environment reward.}
\label{fig:transfer}
\end{figure}

While the actions performed by the simulator policy can fail to accomplish the task on the robot, the sequence of states visited by the agent in simulation represents its task under limited variation in the system dynamics. 
We propose \method~(\meth), 
which instead of directly adapting the simulation policy, guides the exploration for both systems towards mutually aligned state distributions via auxiliary rewards. The method is displayed in Figure \ref{fig:transfer} and employs an adversarial approach to train policies with additional rewards based on confusing a discriminator with respect to the originating system for state sequences visited by the agents.
By guiding the target agent on the robot towards states that the potentially more proficient source agent visits in simulation, we can accelerate training. 
In addition to aligning the robot policy to adapt to progress in simulation, we extend the approach to mutually align both systems which can be beneficial as the agent in simulation will be driven to explore better trajectories from states visited by the real-world policy.

We evaluate the method developed on a set of common reinforcement learning benchmark tasks \citep{duan2016benchmarking} to transfer between simulations with differences in system parameters such as density, dampening and friction. Furthermore, we extend the experiments to address additional challenges relevant in the context of real platforms such as sparse and uninformative rewards. The final experiments investigate transfer between different simulation engines with unknown discrepancies as a stronger proxy for real robot experiments. 

We demonstrate that auxiliary rewards, which guide the exploration on the target platform, improve performance in environments with sparse rewards and can even guide the agent if only uninformative or no environment rewards at all are given for the target agent.
Furthermore, the approach proves to be capable of guiding training given scenarios with significant discrepancies between the system dynamics when direct transfer or fine tuning approaches fail as investigated in Section \ref{sec:frameworks}.

\section{Related Work}
Given significant advances when training machine learning models and in particular reinforcement learning policies in simulations, the field of transfer learning has gained increased relevance in recent years.
In some cases the simulator-given environment represents the final application environment like e.g. when solving computer games, such as various Atari games \citep{mnih2015human} and Ms. Pacman \citep{VanSeijen2017}, as well as board games, such as Chess \citep{campbell2002deep} and Go \citep{Silver2016}.
In most cases however, the direct application of models trained in simulation to real world tasks results in significantly decreased performance. This can be due to two possible types of discrepancies: different observation model or different system dynamics.

The problem of different observation distribution has been addressed generally in the framework of unsupervised domain adaptation \citep{Ganin2016,Long0J16a} as well as with particular application to robotics \citep{tzeng2015towards,stadie2017}.
\citet{rusu2016sim} tackles the transfer task by reusing features learned for simulation and focusing on learning residual representations needed to adapt to the real world task.
Furthermore, recent work addresses the task via visual domain randomization, showing that exact simulation might not be needed when instead training for various variations of textures and colors \citep{tobin2017domain,sadeghi2016cad} without any training data required from the final application system.

The second challenge is based on differences of the system dynamics of simulator and real task, which results in different optimal policies even when given the same observation model, is the focus of our work. \citet{christiano2016transfer} 
address the challenge by learning an inverse dynamics model for the real platform to determine the correct real world action based on the trajectory in simulation.
\citet{gupta2017learning} learn invariant feature spaces which are employed to transfer task knowledge even between systematically different robot architectures. However, the approach relies on prior information through proxy tasks to determine the invariant spaces. Another approach is given by \citet{rajeswaran2016epopt}, who train robust policies based on sampling from an  ensemble of simulated source domains while focusing on simulations with reduced performance. Additional methods from \citep{abbeel2006using,MarcoBHS0ST17} focus on exploiting inaccurate simulations to improve real world performance with the latter employing bayesian optimisation to improve the split between simulator and real world experiments.

Our method builds on a different idea of transfer learning which has its roots in imitation learning where we align the distributions over visited states between two agents. The alignment procedure is realized by training in an adversarial framework \citep{goodfellow2014generative} to confuse a discriminator that itself is learning to classify states based on which system they originated in. The approach is conceptually similar to adversarial imitation learning methods \citep{ho2016generative,stadie2017,hausman2017multi} which train for one-sided alignment between the agent's policy and a set of demonstration trajectories. Instead our method trains both -- simulator and robot -- policies with an auxiliary reward based on the mutual alignment of visited state distributions between both systems as described in greater detail in Section \ref{sec:matl}.

\begin{figure}
    \centering
    \begin{subfigure}[b]{0.3\textwidth}
        \includegraphics[width=\textwidth]{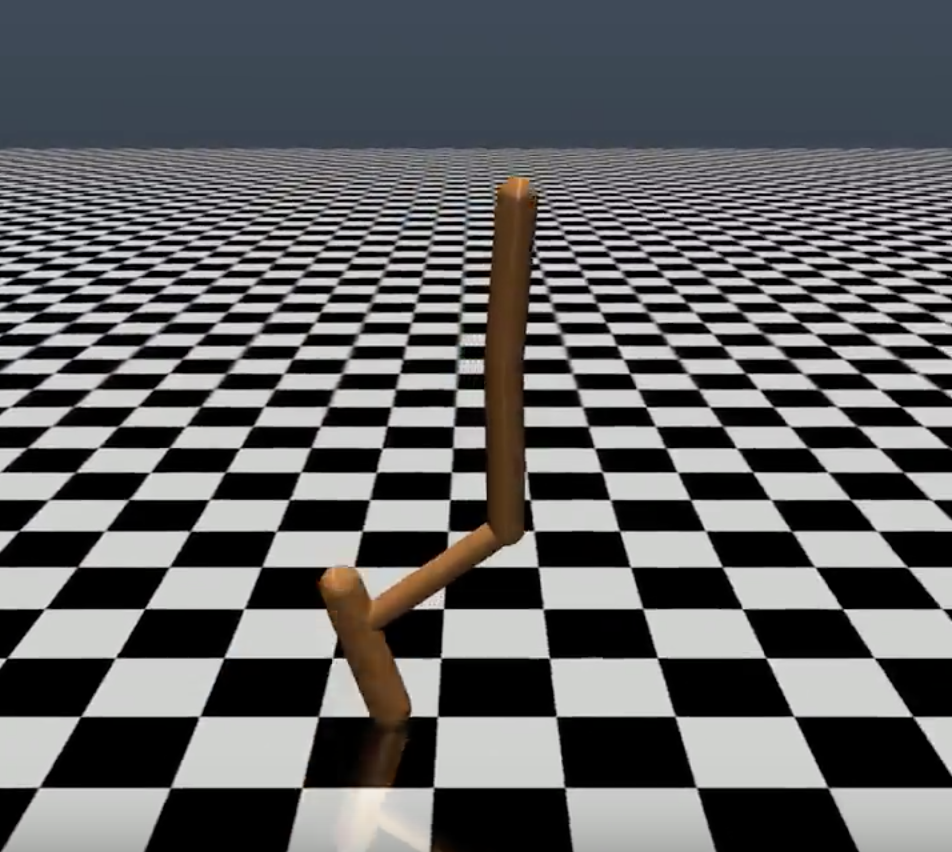}
        \caption{OpenAI gym - MuJoCo}
        \label{fig:muj_hopper_sims}
    \end{subfigure}
    \quad 
    \begin{subfigure}[b]{0.3\textwidth}
        \includegraphics[width=\textwidth]{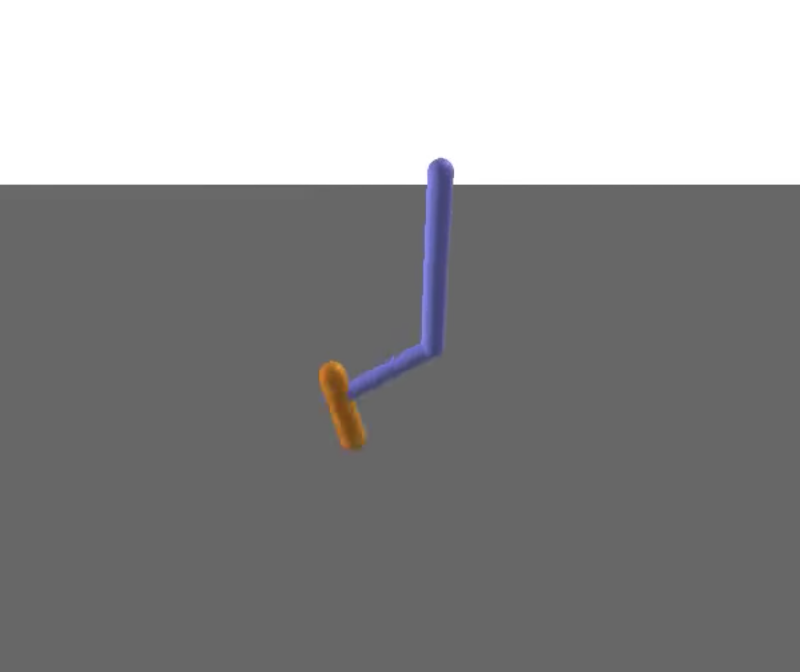}
        \caption{DartEnv - DART}
        \label{fig:dart_hopper_sims}
    \end{subfigure}

    \caption{Hopper2D task build atop MuJoCo \cite{todorov2012mujoco} and DART \cite{dartsim} simulation engines. Both reinforcement learning environments are provided by OpenAI gym \cite{brockman2016openai} and DartEnv \cite{dartenv} respectively. The environments have the same state and action spaces - however differ in the underlying system dynamics and in particular contact modelling. }\label{fig:sims_hopper}
\end{figure}

\section{Method}
\label{sec:method}


We consider a simulation to real robot transfer learning scenario for reinforcement learning with the setup of two agents acting in the respective source and target environments, each represented as Markov Decision Process (MDP). State space $s \in \mathcal{S}$ and action space $a \in \mathcal{A}$ are equal in both environments. However, the difference lies in the underlying system dynamics represented as $p_{S}(s_{t+1}|s_t,a_t)$ for the simulator and $p_{R}(s_{t+1}|s_t,a_t)$ for the real platform. The reward functions  $r_{S}(s_t,a_t)$ and $r_{R}(s_t,a_t)$ which both agents optimize, can be the same for both environments, but can also differ as it is possible that we do not have access to the full reward function in the real world. However, both agents are intended to solve is the same task.
Both agents, simulation/source and robot/target, act according to their respective stochastic policies $\pi_{\theta}(a_t|s_t)$ and $\pi_{\phi}(a_t|s_t)$, which are parameterized as neural networks by $\theta$ and $\phi$.
In the following sections the terms simulator and robot, or respectively source and target are used interchangeably.

\subsection{\method}\label{sec:matl}
\method~(\meth) leverages information gained while training in simulation to improve performance on a target robot with potentially different system dynamics via the introduction of auxiliary reward functions to guide both systems to visit similar states or state sequences. The approach is visualized in Figure \ref{fig:transfer}. 
When the simulator policy cannot be directly applied due to systematic differences, the simulation trajectories still contain information about how to solve the task on a different system given limited variation in the underlying system dynamics. 

The method trains policies with auxiliary alignment rewards by running reinforcement learning algorithms simultaneously on both systems, here Trust Region Policy Optimization (TRPO) \citep{schulman2015trust}. As training in simulation can potentially be performed orders of magnitude faster, we run the updates for the simulator policy at $M$ times higher rate with only the environment reward being used for these updates.
Both policies, $\pi_{\theta}(a|s)$ and $\pi_{\phi}(a|s)$, are trained via TRPO with the method's gradient step computed following Equation \ref{eq:trpo} based on their respective reward signals $r(s,a)$.

\begin{equation}\label{eq:trpo}
    \mathbb{E}_{\pi,p}[\nabla~ \log \pi (a_t|s_t)\:r(s_t,a_t)]
\end{equation}

Auxiliary rewards for the alignment process are generated in an adversarial setting with the objective to confuse the discriminator $D_\omega$ which is trained to classify the system of origin for state sequences $\zeta_{t}$ from simulation and robot. As displayed in Equation \ref{eq:disc}, state sequences can be subsampled to ensure significant change between successive states.
In addition to aligning the robot policy to adapt to progress in simulation, the reciprocal alignment of the simulation policy can be beneficial as the agent in simulation will be driven to explore better behaviours from states visited by the robot agent. 

The discriminator $D_\omega$ is parameterized as neural network with weights $\omega$ and trained according to Equation \ref{eq:disc} to classify the originating system (simulation-robot) of fixed length state sequences $\zeta_t$.

\begin{eqnarray}
    \mathcal{L_D} = \mathbb{E}_{\pi_{\theta}}[\log(D_\omega (\zeta_{t}))] + \mathbb{E}_{\pi_{\phi}}[ \log(1-D_\omega (\zeta_{t}))]\label{eq:disc}\\
     \zeta_{t} = s_{t}, s_{t+k}, s_{t+2k},..., s_{t+nk} \text{~~~~with~} n \in 
     \mathbb{N}_{\pmbb{0}} ; k \in \mathbb{N}   \nonumber  
\end{eqnarray}

Additionally to the simulator and robot environment feedback, respectively $r_{S}(s_t,a_t)$ and $r_{R}(s_t,a_t)$, the full reward includes the auxiliary rewards $\rho_{{S}}$ and $\rho_{{R}}$, for simulator policy and robot policy, as given in Equation \ref{eq:rewards}. By training both policies towards mutual alignment, not only does the robot policy learn from the progress in simulation but also is the simulator policy pushed to explore better behaviour for states visited in the robot environment. 
The specific formulation in Equations \ref{eq:rewards_simaux} and \ref{eq:rewards_realaux} builds upon the idea of maximizing the confusion loss formulation of the GAN framework \citep{goodfellow2014generative}
, which was found empirically to be better suited for the transfer task than the original min-max formulation. The confusion objective for adversarial training addresses a principal shortcoming of the original formulation by which the gradients for the generating module (represented here by the policies) vanish when the discriminator performance is maximized. Hereinafter, the subscripts $R$ and $S$ stand for reference to robot and simulator systems respectively.


\begin{eqnarray}
    r(s_t,a_t) &=& \begin{cases}
 r_{R}(s_t,a_t) + \lambda \: \rho_{R}(s_t)  &\text{: robot agent}\\
 r_{S}(s_t,a_t) + \lambda \: \rho_{S}(s_t)  &\text{: simulator agent}
\end{cases}
 \label{eq:rewards}\\
    \rho_{{S}}(s_t) &=& -\log(D_\omega (\zeta_{t}))\label{eq:rewards_simaux}\\
    \rho_{{R}}(s_t) &=& \log(D_\omega (\zeta_{t}))\label{eq:rewards_realaux}
\end{eqnarray}

In conclusion, the full gradient steps for TRPO in the MATL framework are obtained by combining Equations and \ref{eq:trpo}, \ref{eq:rewards}, \ref{eq:rewards_simaux} and \ref{eq:rewards_realaux}. Both updates for the simulator and robot policies are given in Equations \ref{eq:main_obj_sim} and \ref{eq:main_obj_real} respectively, with the complete training procedure in Algorithm \ref{alg:matl}.

\begin{eqnarray}
    &&\mathbb{E}_{\pi_{\theta}}[\nabla_\theta~ \log \pi_{\theta} (a_t|s_t)\:(r_{S}(s_t,a_t)-\lambda \:\log(D_\omega (\zeta_{t})) )]\label{eq:main_obj_sim}\\ 
    &&\mathbb{E}_{\pi_{\phi}}[\nabla_\phi~ \log \pi_{\phi} (a_t|s_t)\: (r_{R}(s_t,a_t)+\lambda \:\log(D_\omega (\zeta_{t}))]\label{eq:main_obj_real} 
\end{eqnarray}


\begin{algorithm}[h]
\SetAlgoLined
 \SetKwInOut{Input}{Input}
 \SetKwInOut{Output}{Output}
 \Input{environments MDP$_{R,S}$, alignment weight $\lambda$, iterations outer loop $N$ and inner loop $M$, rollout horizon $T$}
 \Output{target policy $\pi_\phi$}
 $\pi_\theta, \pi_\phi$, $D_\omega \leftarrow$ \text{initialize}\
  
 \For{$i ~\leftarrow \ 1$ \KwTo $N$}{
   $(s_{0..T},a_{0..T},r_{0..T})_S \leftarrow \text{rollout} ~\pi_\theta ~\text{on MDP}_S$\
   
  $(s_{0..T},a_{0..T},r_{0..T})_R \leftarrow \text{rollout} ~\pi_\phi ~\text{on MDP}_R$\
  
  $D_\omega \leftarrow$ \text{gradient update following Eq. \ref{eq:disc}}\
  
  $\pi_\theta,\pi_\phi \leftarrow$\text{TRPO updates following Eqs. \ref{eq:main_obj_sim}~\&~\ref{eq:main_obj_real}}\
  
  \For{$j ~\leftarrow \ 1$ \KwTo $M$}{
  
    $(s_{0..T},a_{0..T},r_{0..T})_S \leftarrow \text{rollout} ~\pi_\theta ~\text{on MDP}_S$\
    
    $\pi_\theta \leftarrow$\text{TRPO update following Eq. \ref{eq:main_obj_sim} with $\rho_{{S}}(s_t,a_t)=0 ~\forall~ t \in T$}\
  
  }
 }
 \caption{\method}
 \label{alg:matl}
\end{algorithm}

\section{Experiments}\label{sec:exps}

We evaluate \meth~on transfer learning scenarios based on various common reinforcement learning tasks including environments taken from rllab \citep{duan2016benchmarking}, OpenAI gym \citep{brockman2016openai} and DartEnv \citep{dartenv}. 
For Sections \ref{sec:sparse} to \ref{sec:only align} we use variations of the same simulation environment with strongly varied parameters for the system dynamics. To evaluate our approach for application with significantly inaccurate knowledge of system dynamics, we severely alter the parameters for joint dampening, friction and densities between both systems. The reader is referred to the additional documentation\footnote{\label{note1}Additional information to environment and algorithm parameterization can be found under sites.google.com/view/matl} for detailed information about algorithm and environment parameterization. 
To extend the evaluation, we address transfer learning between two different simulation engines, MuJoCo \citep{todorov2012mujoco} and DART \citep{dartsim}, in Section \ref{sec:frameworks}.

The evaluation focuses six different approaches in the following experiments:
\begin{itemize}
    \item independent - Independent training of the robot policy without auxiliary rewards.
    \item direct\_transfer - Direct application of the simulator policy on the real platform.
    \item fine\_tuning - Transfer of the fully trained simulator policy and subsequent fine tuning based only on robot environment rewards.
    \item MATLu - Unilateral alignment training with environment rewards but without auxiliary reward for the simulation policy.
    \item MATL - Mutual alignment training with environment rewards and both auxiliary rewards.
    \item MATLf - Combining MATL with fine tuning. Training via MATL but starting from a transferred fully trained simulator policy.  
\end{itemize}

All diagrams focus on the number of real world iterations as additional iterations in simulation can be obtained at significantly higher rate and (in comparison) negligible effort. The performance for each experiment is normalized between zero and one.

\subsection{Guiding Questions}
\begin{itemize}
    \item Are MATL's auxiliary alignment rewards suited to guide exploration in environments with sparse rewards? (Section \ref{sec:sparse}) 
    \item Do the auxiliary rewards provide enough feedback on how to solve a task in a situation where the real world agent has only access to an uninformative reward (such as a reward for not falling)? (Section \ref{sec:uninformative})
    \item Can we succeed in training towards a task with no reward given in the target environment other than the auxiliary rewards (additionally deactivating all environment rewards in the target environment including the reward for staying alive)? (Section \ref{sec:only align}) 
    \item How important is the mutual alignment vs a unilateral alignment of just the target policy? (Section \ref{sec:sparse} to Section \ref{sec:frameworks})
    \item How well does the approach handle transfer between different simulation frameworks with unknown differences in system dynamics? (Section \ref{sec:frameworks} )

\end{itemize}

\subsection{Sparse Rewards} \label{sec:sparse}
The first section of our evaluation focuses on the application of \meth~to environments with sparse rewards. In these experiments, both environments - simulation and real - only have access to sparse feedback which is given when the agent is situated within a distance $\epsilon$ to the target position\textsuperscript{\ref{note1}}  instead of a dense, distance based reward. 

Sparse feedback renders these situations more complex than scenarios with dense rewards 
as the agents only seldomly get feedback and learning and progress is delayed. The auxiliary rewards given by MATL however are non-sparse and can help guide the real world agent to solve the task, given it has learned to solve the task in simulation.
For these simpler tasks it was found empirically sufficient to apply the discriminator to single states.


\begin{figure}[h]
    \centering
    \begin{subfigure}[b]{0.32\textwidth}
        \includegraphics[width=\textwidth]{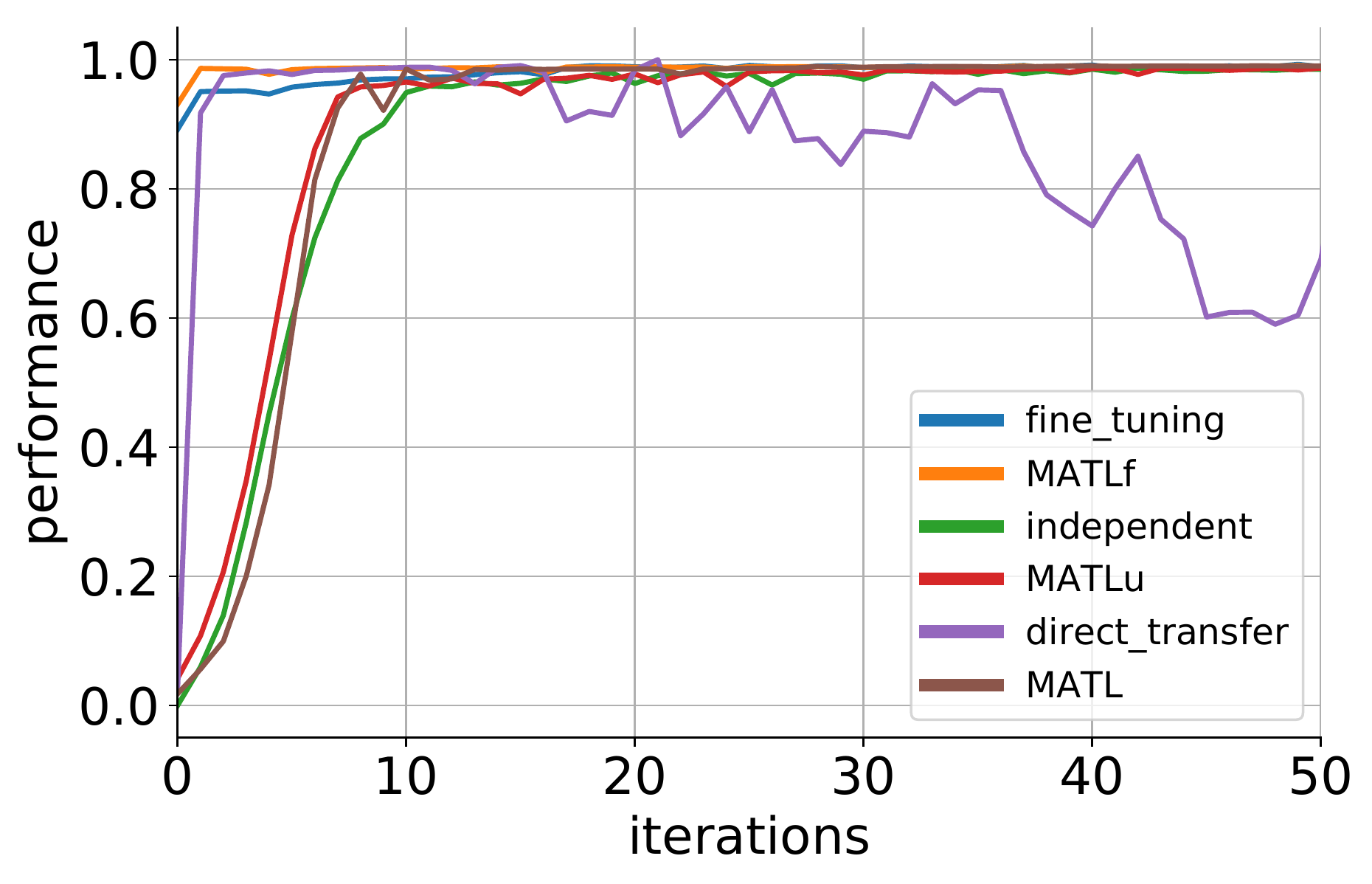}
        \caption{Cartpole}
        \label{fig:cartpole}
    \end{subfigure}
    \begin{subfigure}[b]{0.32\textwidth}
        \includegraphics[width=\textwidth]{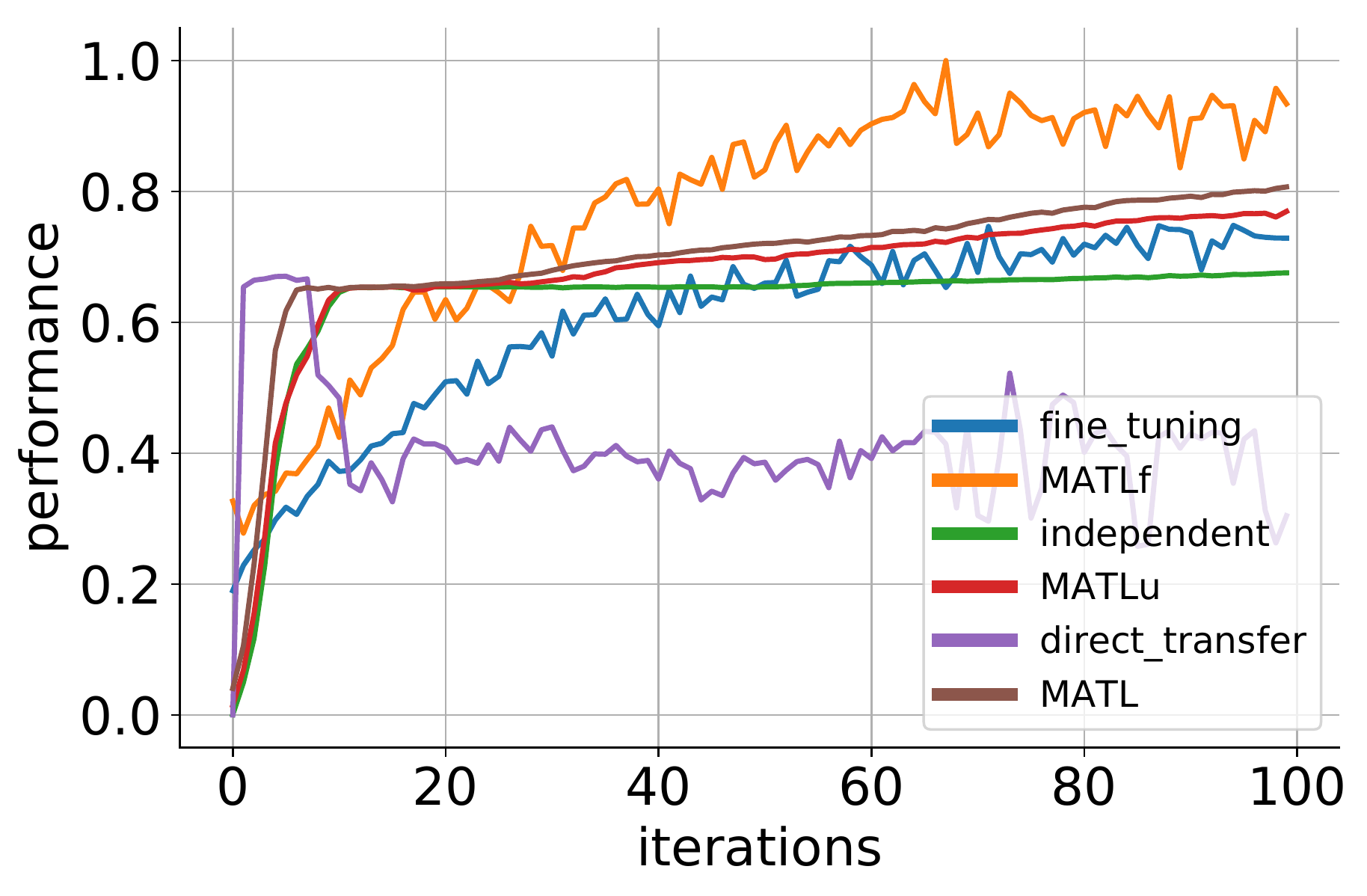}
        \caption{Cartpole Swingup}
        \label{fig:swingup}
    \end{subfigure}
    \begin{subfigure}[b]{0.32\textwidth}
        \includegraphics[width=\textwidth]{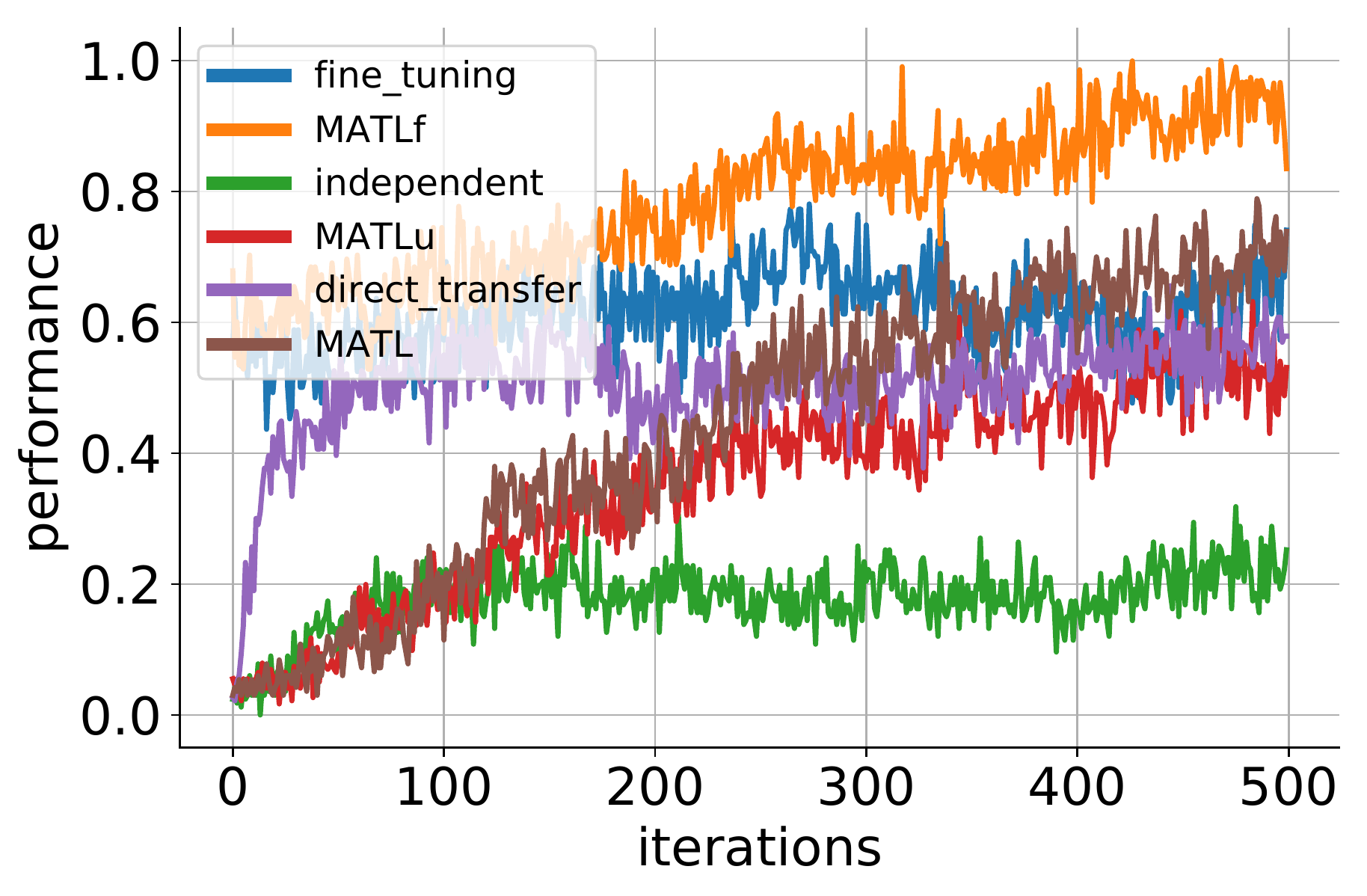}
        \caption{Reacher2D}
        \label{fig:swingup}
    \end{subfigure}
    \caption{Transfer learning with sparse rewards in both environments. The dense auxiliary rewards from \meth~help to accelerate training. While fine tuning already learns to perform optimally after a few iterations, MATLf accelerates training even further.   }\label{fig:sparse}
\end{figure}

The direct transfer of the simulation policy is unsuited for application on the target system in both tasks as displayed in Figure \ref{fig:sparse}. While independent training already learns fast and fine tuning adapts even better, the best performance is achieved by combining \meth~with fine tuning.

\subsection{Uninformative Rewards}\label{sec:uninformative}
We furthermore investigate scenarios limited to an uninformative reward for locomotion tasks.
In these cases, the only remaining component is a cost for falling in the target environment. The agent in simulation is still provided with access to the full reward including a forward-guiding component. 
We evaluate these scenarios based on how well the agent learns to move forward and therefore the capability of MATL to guide to forward motion. The performance is now given as metric based on the average final distance of the agent in the direction of locomotion.

Similar to \citet{stadie2017} we exploit in this section the sequential structure of these tasks and apply the discriminator to state pairs of time-steps t and t+4, which has been shown to work well across a variety of different tasks.


\begin{figure}[h]
    \centering
    \begin{subfigure}[b]{0.35\textwidth}
        \includegraphics[width=\textwidth]{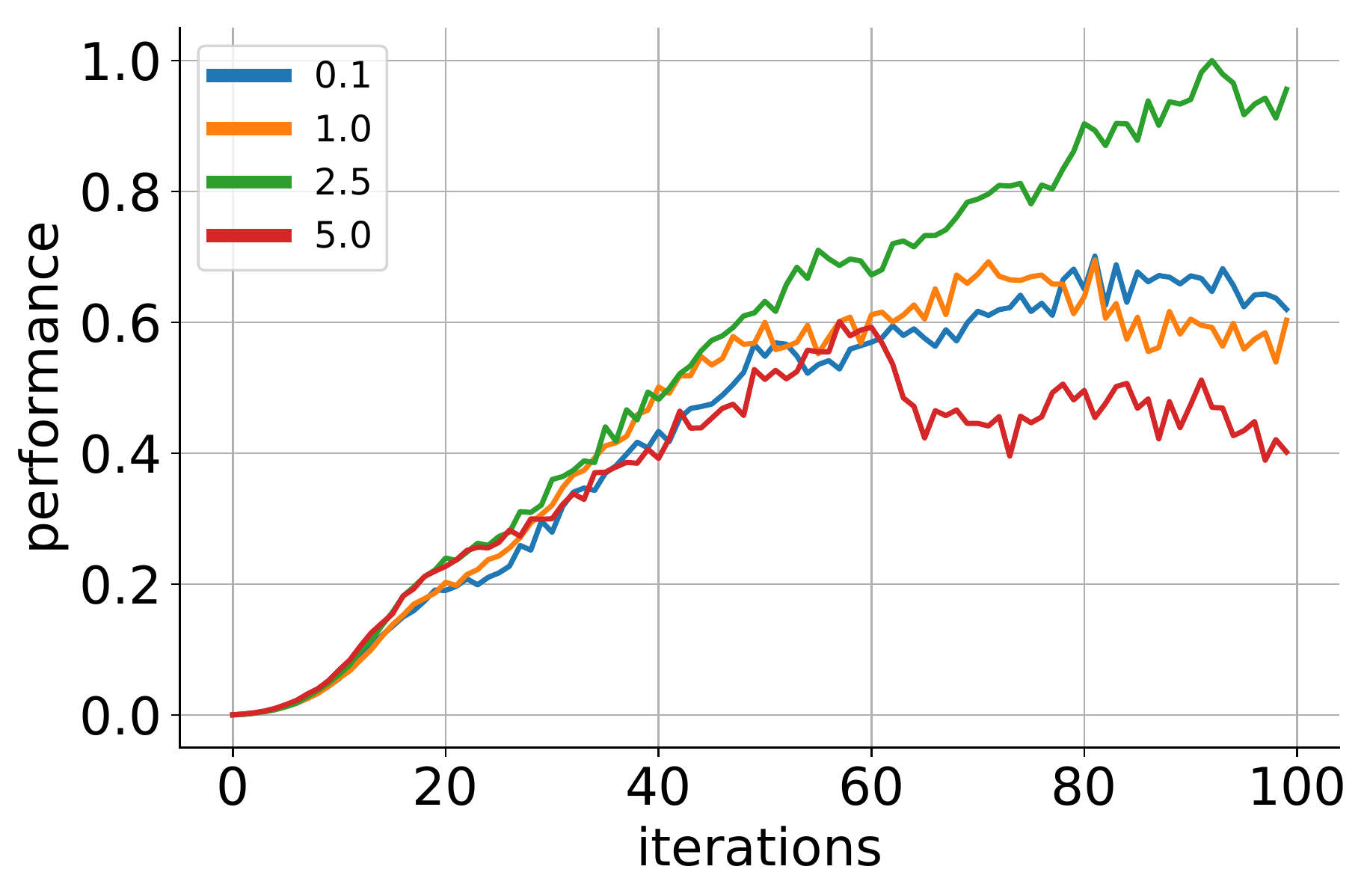}
        \caption{Hopper2D - Alignment weight}
        \label{fig:hopper_align}
    \end{subfigure}
    ~ 
    \begin{subfigure}[b]{0.35\textwidth}
        \includegraphics[width=\textwidth]{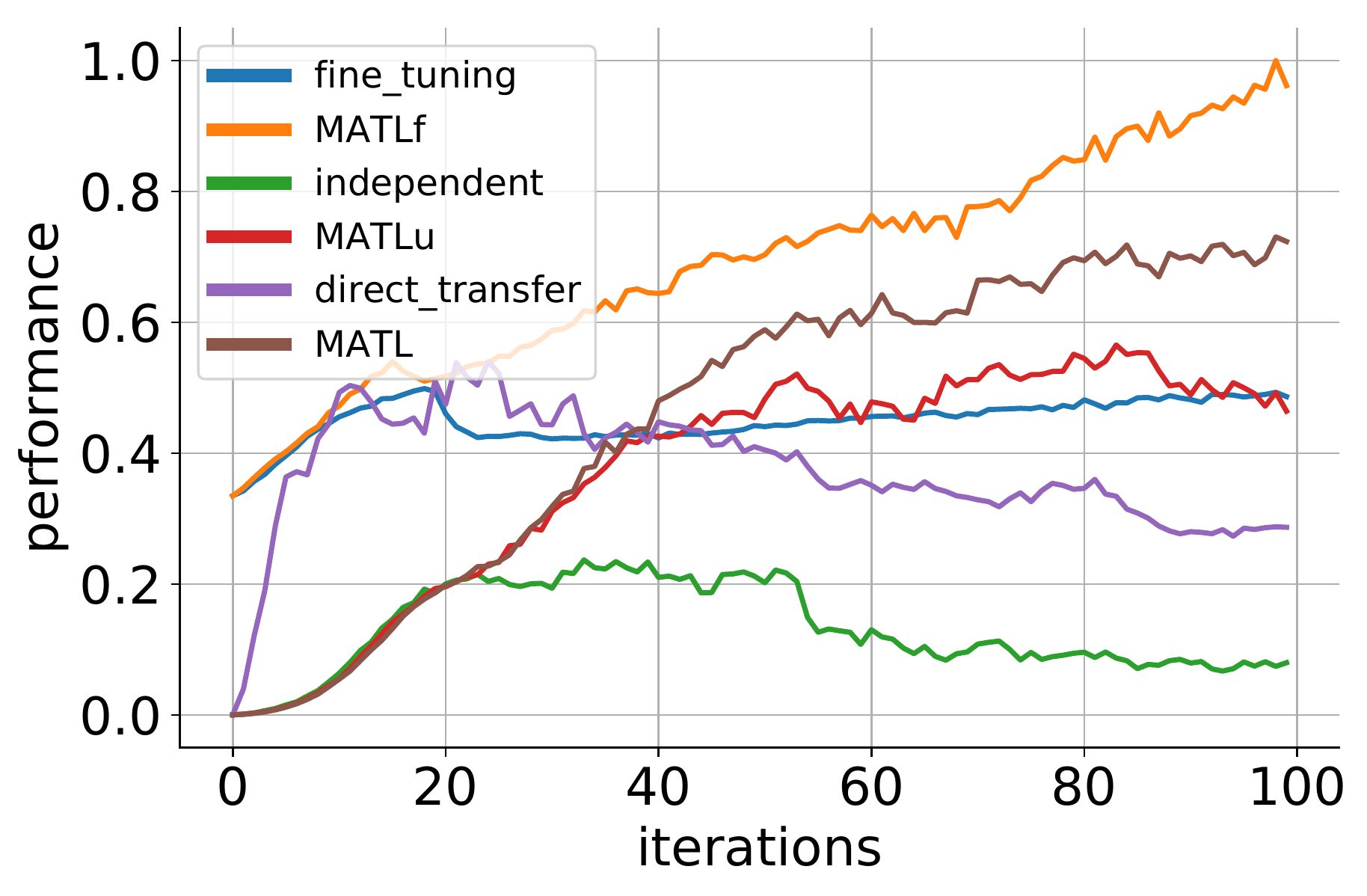}
        \caption{Hopper2D - Methods}
        \label{fig:hopper_perform}
    \end{subfigure}
    \caption{Transfer learning - Hopper2D task with uninformative rewards. Left: Transfer learning performance in dependency of alignment weight. The context of potentially conflicting rewards renders the hyperparameter choice more critical for efficient transfer. Right: Comparison against baselines with MATLf surpassing the other methods. }\label{fig:uninformative}
\end{figure}

The most robust policy is simply standing still as moving forward increases the risk of falling. This situation renders the task more complex as the auxiliary reward has to overcome a potentially conflicting reward signal. Figure \ref{fig:hopper_align} displays that the alignment parameter $\lambda$ is highly relevant to encourage stable walking behaviour. The conflict between the reward components results in a more conservative, ankle-based running style as can be found in the videos as part of the additional material\textsuperscript{\ref{note1}}.

Figure \ref{fig:uninformative} shows that independent training results in very limited motion as reward signal encourages the agent to stand still. Fine tuning as well as direct transfer of the simulation policy perform slightly better, but are significantly surpassed by all versions of \meth.

\subsection{Without Robot Environment Reward} \label{sec:only align}
To evaluate if transfer is possible without any environment reward in the target environment we run a set of experiments with only auxiliary reward based updates for the robot policy. The changes in parameterization between simulation and robot environments are equal to the settings for Sections \ref{sec:sparse} and \ref{sec:uninformative}.

Contrary to the other sections, the performance is given for these experiments as ratio of the maximum performance achieved in the same environment with available environment reward from Section \ref{sec:sparse} and \ref{sec:uninformative}. This metric is chosen to evaluate the relative performance reduction for all methods when access to environment rewards is prevented.


\begin{figure}[h]
    \centering
        \begin{subfigure}[b]{0.32\textwidth}
        \includegraphics[width=\textwidth]{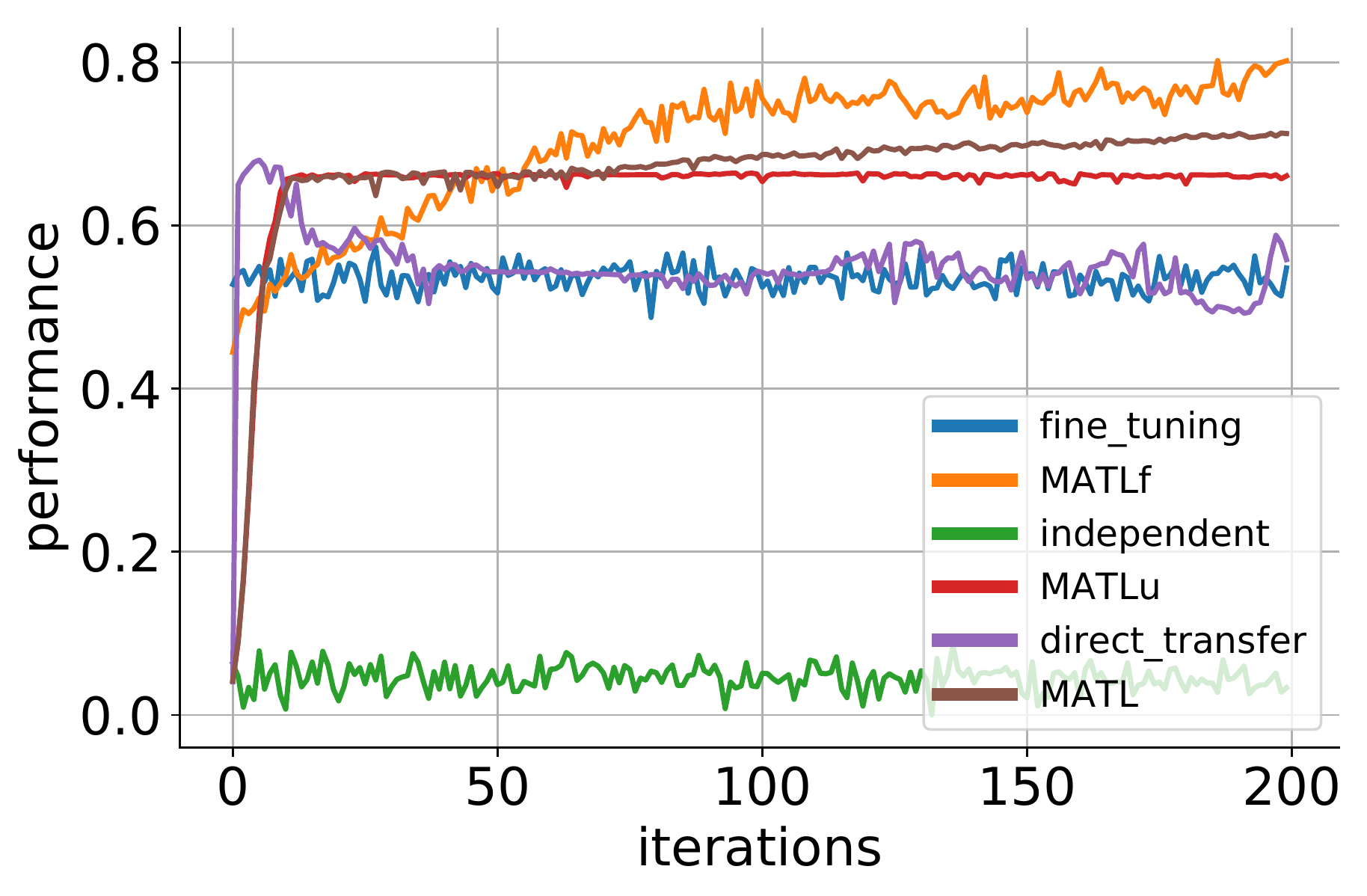}
        \caption{Cartpole Swingup}
        \label{fig:swingup_no_env}
    \end{subfigure}
    \begin{subfigure}[b]{0.32\textwidth}
        \includegraphics[width=\textwidth]{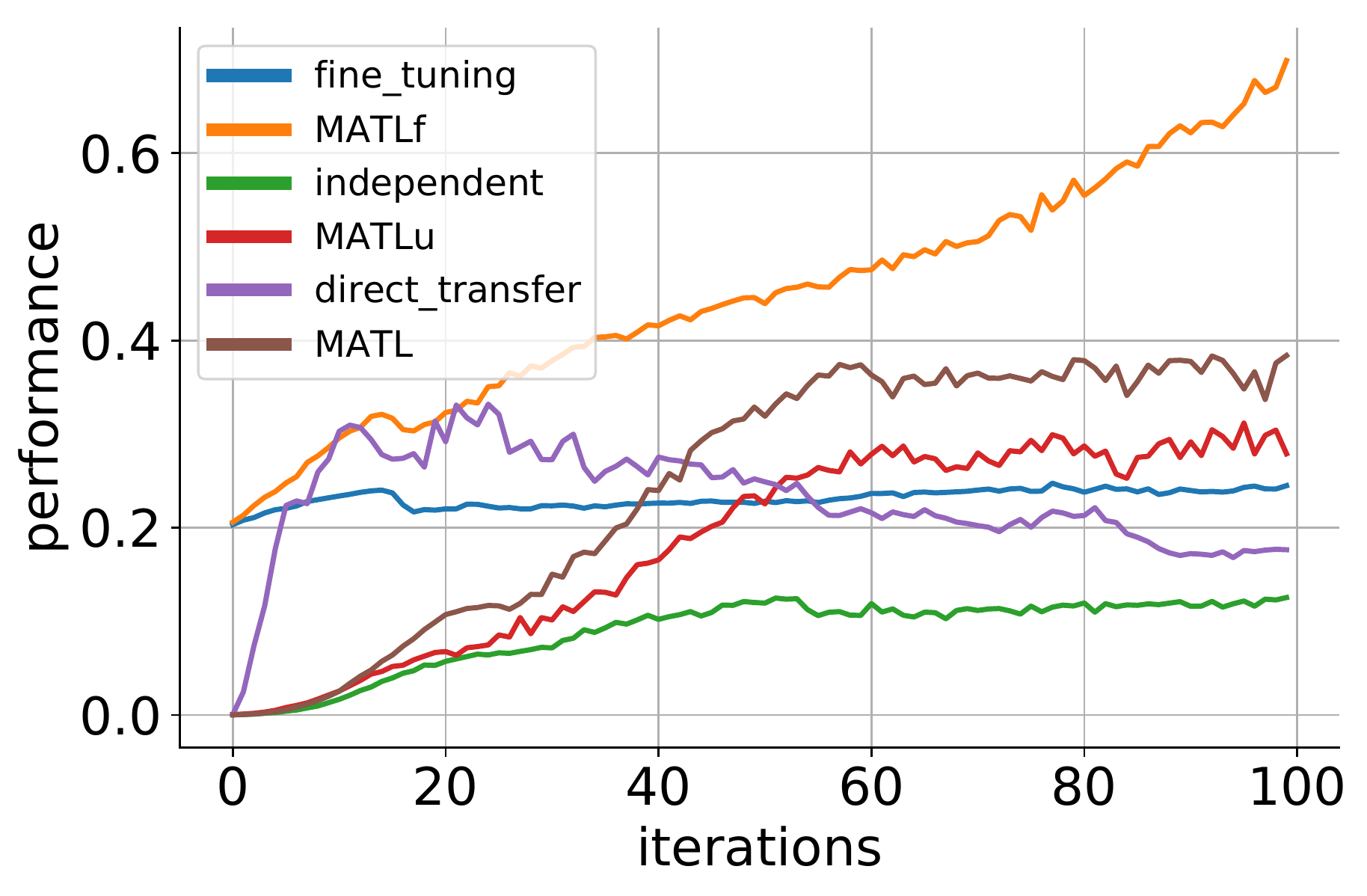}
        \caption{Hopper2D}
        \label{fig:hopper_no_env}
    \end{subfigure}
    \begin{subfigure}[b]{0.32\textwidth}
        \includegraphics[width=\textwidth]{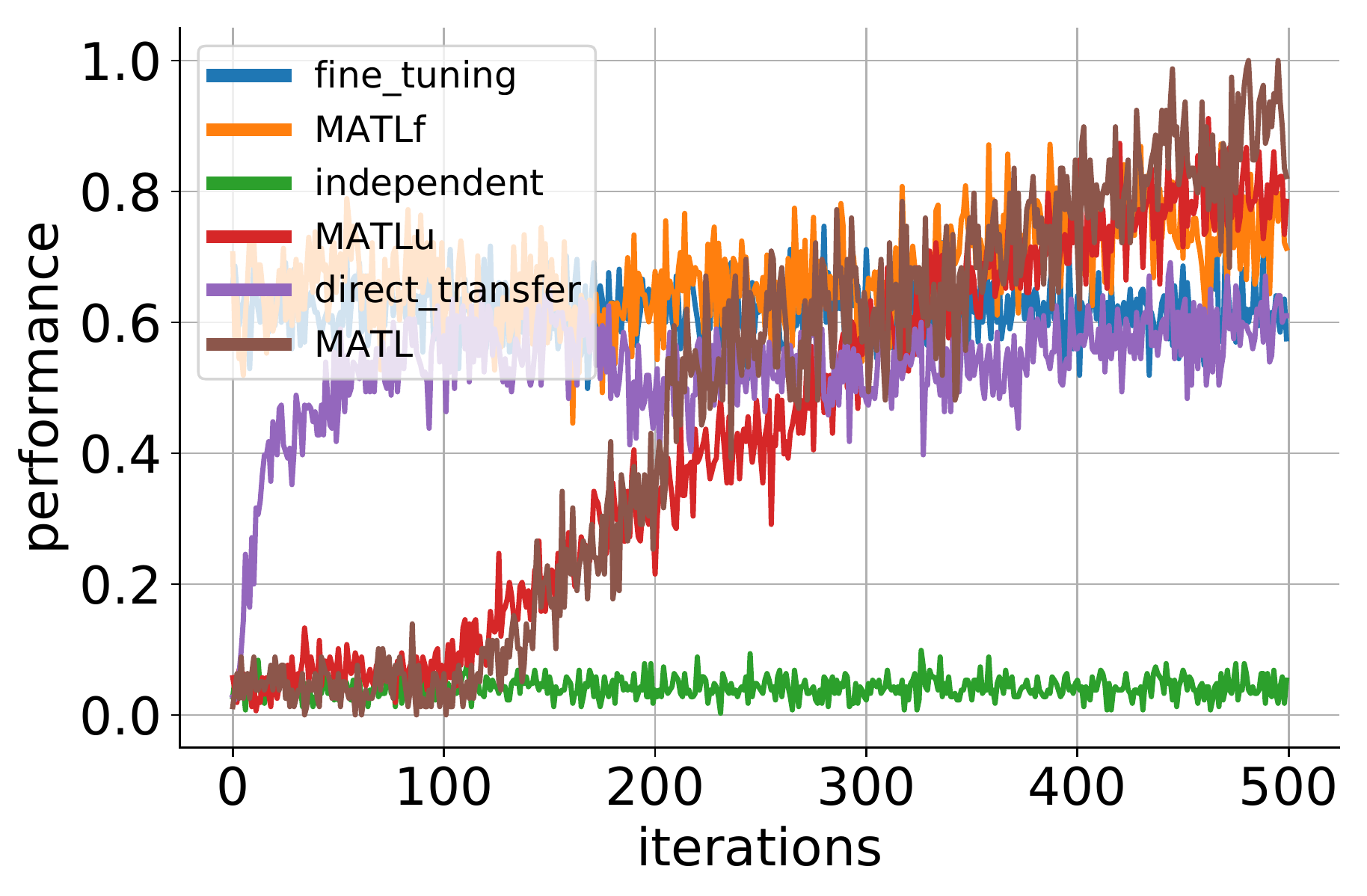}
        \caption{Reacher2D}
        \label{fig:hopper_no_env}
    \end{subfigure}
    \caption{Transfer learning - Scenario without environment reward on the real platform }\label{fig:no_env}
\end{figure}

In all scenarios, MATLf -- the combination of fine tuning and auxiliary rewards -- outperforms the other methods. This fine-tuning of a simulator-trained policy via MATL improves from a well-suited initialization, often exceeding other versions of MATL in its maximally achieved reward. As expected, these environments result in no significant progress for methods training only on target rewards such as fine tuning, which keeps the same performance as the pretrained policy.

\subsection{MuJoCo to DART} \label{sec:frameworks}
While the earlier experiments are based on the MuJoCo simulator and varied parameters for different system properties, we extend the evaluation of our approach towards differences between two simulation software packages, namely from MuJoCo \citep{todorov2012mujoco} to DART \citep{dartsim}.
An additional challenge of this transfer experiment is that not only parameter values vary but the underlying algorithms differ in terms of parameter types and in particular contact modeling \citep{dartenv}.

\begin{figure}[h]
    \centering
    \begin{subfigure}[b]{0.35\textwidth}
        \includegraphics[width=\textwidth]{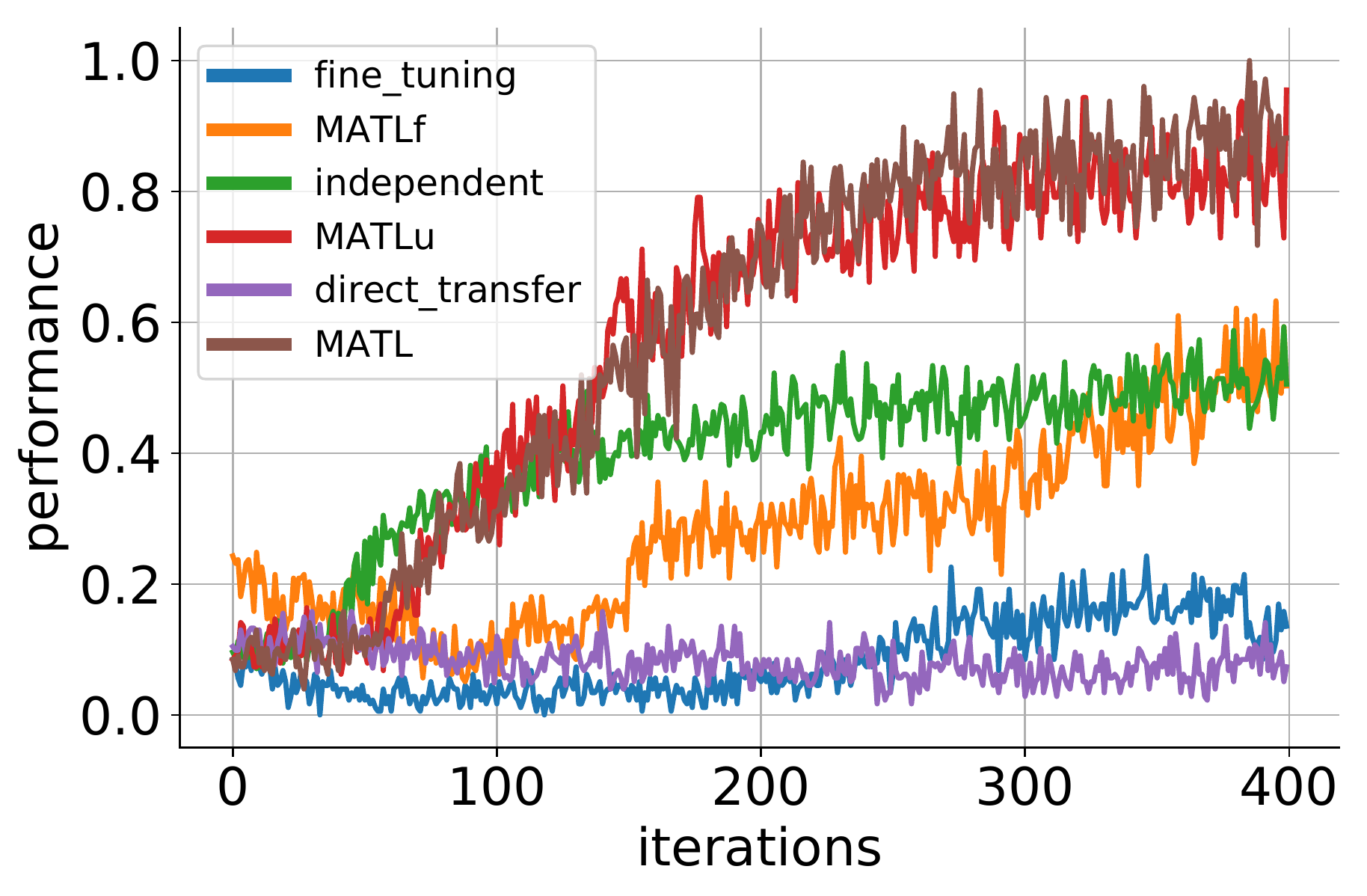}
        \caption{Reacher}
        \label{fig:dart_reacher}
    \end{subfigure}
    ~ 
    \begin{subfigure}[b]{0.35\textwidth}
        \includegraphics[width=\textwidth]{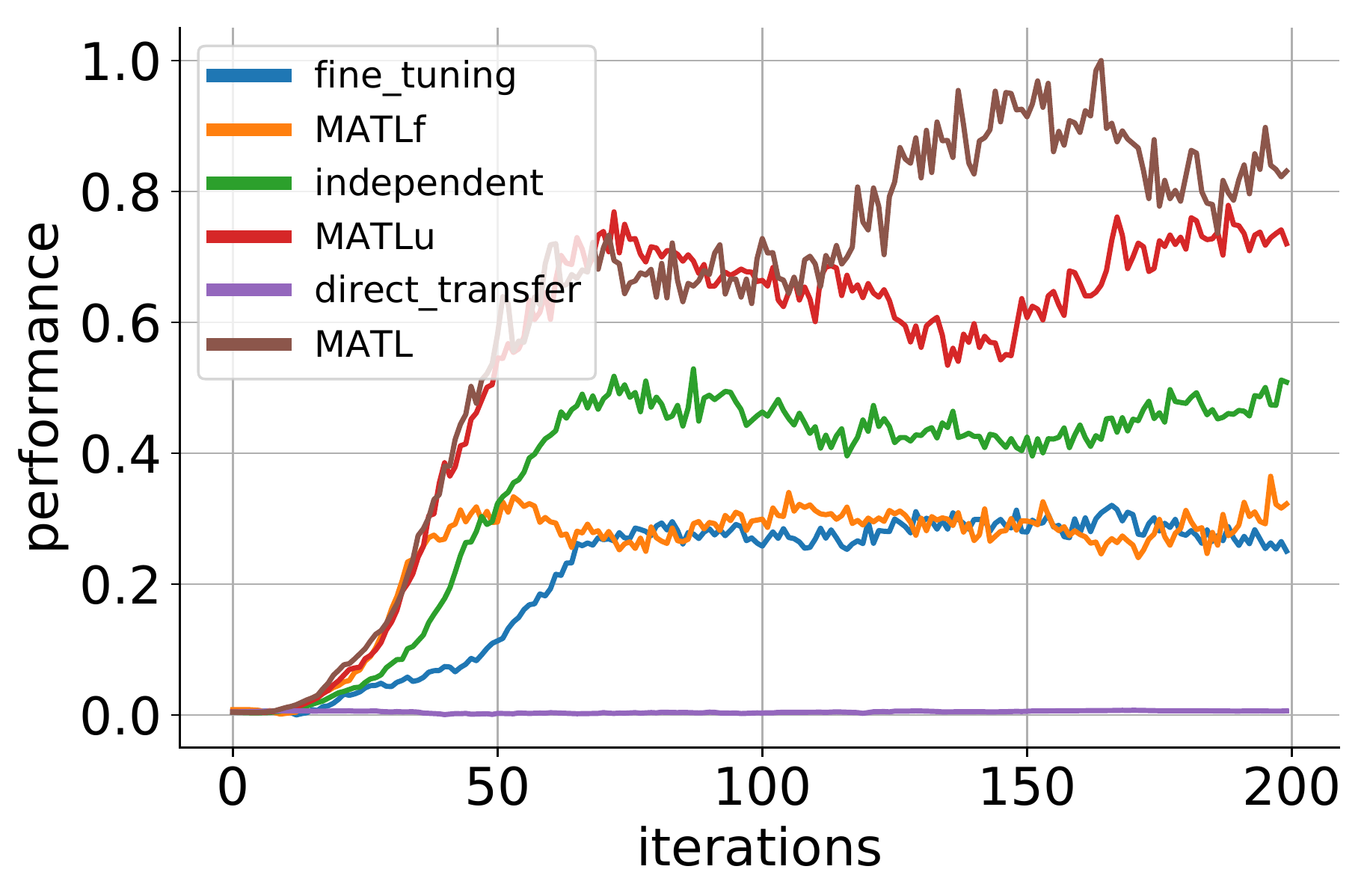}
        \caption{Hopper2D}
        \label{fig:dart_hopper}
    \end{subfigure}

    \caption{Transfer learning - MuJoCo to DART simulator }\label{fig:simulators}
    \vspace{-5mm}
\end{figure}

The confusion loss formulation for the original GAN framework did not result in improvement for these experiments and presented results are based on an adaptation of the Wasserstein GAN (WGAN) loss \citep{arjovsky2017wasserstein}, which is described in greater detail as part of the additional materials.

As the DART environments differ significantly from their MuJoCo counterparts, direct transfer as well as fine tuning result in low performance. In particular, it can be seen that pretraining in one environment results to a low quality initialization and independently training on the target environment surpasses the fine tuning approach. This inherently also leads to low performance for MATLf which builds on the pretrained policy, but does not affect MATLu and \meth.

\section{Discussion}
\meth~has been shown to work under significant differences in system dynamics between source and target platform as demonstrated in Section \ref{sec:exps}, including situations when direct transfer of the simulator policy fails to show good performance. A current shortcoming is the potential instability of the adversarial training framework and connected effort in tuning the hyperparameters.
The alignment weight parameter $\lambda$ is of particular importance in the context of potentially conflicting rewards as is represented by the uninformative rewards in Section \ref{sec:uninformative}. The weight has to be increased above the value of 0.1 which is used for most other experiments as the safety based environment rewards will prevent exploration.

Different simulation engines, as given in Section \ref{sec:frameworks}, provide a particular challenge for the transfer learning methods.
We show that in these cases, the simulation based policy can overfit and result in providing an unsuitable initialization for fine tuning which performs worse during training than standard random initializations. Nevertheless, \meth~demonstrably accelerates training under these conditions. 
Mutual alignment additionally increases performance consistently across all locomotion tasks (Sections \ref{sec:sparse} and \ref{sec:only align}) while being commensurate with unilateral alignment on the tasks with sparse rewards given in Section \ref{sec:sparse}.

While distribution based alignment has been demonstrated to work well in the experiments in Section \ref{sec:exps}, evaluations based on the straightforward approach of direct alignment between the states along trajectories of each system only lead to limited performance improvements. Auxiliary rewards based on state-wise alignment of simulator and real world trajectories perform adequate mostly in low dimensional tasks with near-deterministic behaviour of the completely trained agent. 
The task of moving a ball towards a goal in 2D by applying forces in 2 directions serves as an example for this kind of scenario, where the optimal trajectory is - independent of initialization -  a straight line between start and target.

\section{Conclusions and Future Directions}

We present an approach for transfer learning under discrepancies in system dynamics for simulation to robot transfer. The approach relies on parallel training of both agents and employs auxiliary rewards to align their respective distributions over visited states and can be straightforwardly supplemented with ideas based on fine tuning.

Guiding robot exploration via alignment of state distributions between both systems has been shown to be beneficial for accelerating training and potentially lead to better final performance in scenarios with sparse or uninformative rewards for the target platform. 

All experiments included in this paper are concerned with transfer learning between two simulations with either different parameterizations or completely different simulation engines to create situations of misaligned and unknown system dynamics. Future work will address the full simulation to robot transfer scenario. As \meth~employs partially trained policies on the real platform, further development will have to address methods for safe application of RL on real world platforms \citep{achiam2017constrained,kahn2017uncertainty}.
Furthermore, one of the principal challenges going forward is the weighting of auxiliary rewards and original environment rewards in particular in situations where both can lead to conflicting behaviours.


\acknowledgments{The authors would like to acknowledge the support of the UK’s Engineering and Physical Sciences Research Council (EPSRC) through the Doctoral Training Award (DTA) as well as the support of the Hans-Lenze-Foundation.
}

\bibliographystyle{plain}
\bibliography{main}

\end{document}